\documentclass[sigconf]{acmart}
\AtBeginDocument{%
  }
\usepackage{listings}
\usepackage{xcolor}
\usepackage{multirow}
\usepackage{balance}
\copyrightyear{2025}
\acmYear{2025}
\setcopyright{acmlicensed}
\acmConference[CIKM '25]{Proceedings of the 34th ACM International Conference on Information and Knowledge Management}{November 10--14, 2025}{Seoul, Republic of Korea}
\acmBooktitle{Proceedings of the 34th ACM International Conference on Information and Knowledge Management (CIKM '25), November 10--14, 2025, Seoul, Republic of Korea}
\acmDOI{10.1145/3746252.3761549}
\acmISBN{979-8-4007-2040-6/2025/11}
\acmSubmissionID{5379}

\title{Leveraging Generative Models for Real-Time Query-Driven Text Summarization in Large-Scale Web Search}

\begin{document}


% \author{Zeyu Xiong}

% % \authornote{Both authors contributed equally to this research.}
% % \orcid{1234-5678-9012}
% % \author{G.K.M. Tobin}
% % \email{webmaster@marysville-ohio.com}
% \affiliation{%
%   \institution{Baidu Inc.}
%   \city{Beijing}
%   % \state{Ohio}
%   \country{China}
% }

\author{Zeyu Xiong}
\authornote{Equal contribution.}
\affiliation{%
  \institution{Baidu Inc.}
  \city{Beijing}
  \country{China}
}
\email{jerainxiong@gmail.com}

\author{Yixuan Nan}
\authornotemark[1]
\affiliation{%
  \institution{Institute of Information Engineering, Chinese Academy of Sciences}
  \city{Beijing}
  \country{China}
}
\email{nanyixuan@iie.ac.cn}

\author{Li Gao}
\authornote{Corresponding author.}
\affiliation{%
  \institution{Baidu Inc.}
  \city{Beijing}
  \country{China}
}
\email{gaoli.sinh@gmail.com}

\author{Hengzhu Tang}
\affiliation{%
  \institution{Baidu Inc.}
  \city{Beijing}
  \country{China}
}
\email{hengzhutang@gmail.com}

\author{Shuaiqiang Wang}
\affiliation{%
  \institution{Baidu Inc.}
  \city{Beijing}
  \country{China}
}
\email{wangshuaiqiang@baidu.com}

\author{Junfeng Wang}
\affiliation{%
  \institution{Baidu Inc.}
  \city{Beijing}
  \country{China}
}
\email{wangjunfeng@baidu.com}

\author{Dawei Yin}
\affiliation{%
  \institution{Baidu Inc.}
  \city{Beijing}
  \country{China}
}
\email{yindawei@acm.org}
\renewcommand{\shortauthors}{Zeyu Xiong et al.}

\begin{abstract}
In the dynamic landscape of large-scale web search, Query-Driven Text Summarization (QDTS) aims to generate concise and informative summaries from textual documents based on a given query, which is essential for improving user engagement and facilitating rapid decision-making. Traditional extractive summarization models, based primarily on ranking candidate summary segments, have been the dominant approach in industrial applications. However, these approaches suffer from two key limitations: 1) The multi-stage pipeline often introduces cumulative information loss and architectural bottlenecks due to its weakest component; 2) Traditional models lack sufficient semantic understanding of both user queries and documents, particularly when dealing with complex search intents. In this study, we propose a novel framework to pioneer the application of generative models to address real-time QDTS in industrial web search. Our approach integrates large model distillation, supervised fine-tuning, direct preference optimization, and lookahead decoding to transform a lightweight model with only 0.1B parameters into a domain-specialized QDTS expert. Evaluated on multiple industry-relevant metrics, our model outperforms the production baseline and achieves a new state of the art. Furthermore, it demonstrates excellent deployment efficiency, requiring only 334 NVIDIA L20 GPUs to handle \textasciitilde50,000 queries per second under 55~ms average latency per query.
\end{abstract}

\begin{CCSXML}
<ccs2012>
   <concept>
       <concept_id>10002951.10003317.10003347.10003357</concept_id>
       <concept_desc>Information systems~Summarization</concept_desc>
       <concept_significance>500</concept_significance>
       </concept>
 </ccs2012>
\end{CCSXML}

\ccsdesc[500]{Information systems~Summarization}


\keywords{Generative Model, Query-Driven Text Summarization, Web Search}


\maketitle

\section{Introduction}
% In the era of information explosion, users increasingly rely on search engines such as Baidu and Google to quickly access the information they need. To help users make efficient decisions amid massive amounts of data, large-scale industrial search systems not only aim to deliver accurate, timely, and highly relevant results, but also provide text summaries on the search results page to preview content. These summaries play a critical role in influencing user click behavior. They must concisely convey the core content of the retrieved document while aligning closely with the user’s query intent, all within just a few lines~\cite{dang2006duc, park2022qsg, deng2020joint}. This type of query-driven summarization task is also called query-oriented summarization\cite{peng2016high} or query-focused summarization\cite{yao2017recent}, which aims to extract or generate summaries that reflect the core information of a document in response to a given query. 
% Query-Driven Text Summarization (QDTS)，

To enhance user experience and facilitate rapid decision-making, large-scale web search systems such as Google and Baidu are increasingly focusing on Query-Driven Text Summarization (QDTS) — a task aimed at generating concise and informative summaries tailored to user queries~\cite{dang2006duc, park2022qsg, deng2020joint,peng2016high,yao2017recent}.

\begin{figure}[t]
  \centering
  \includegraphics[width=\columnwidth]{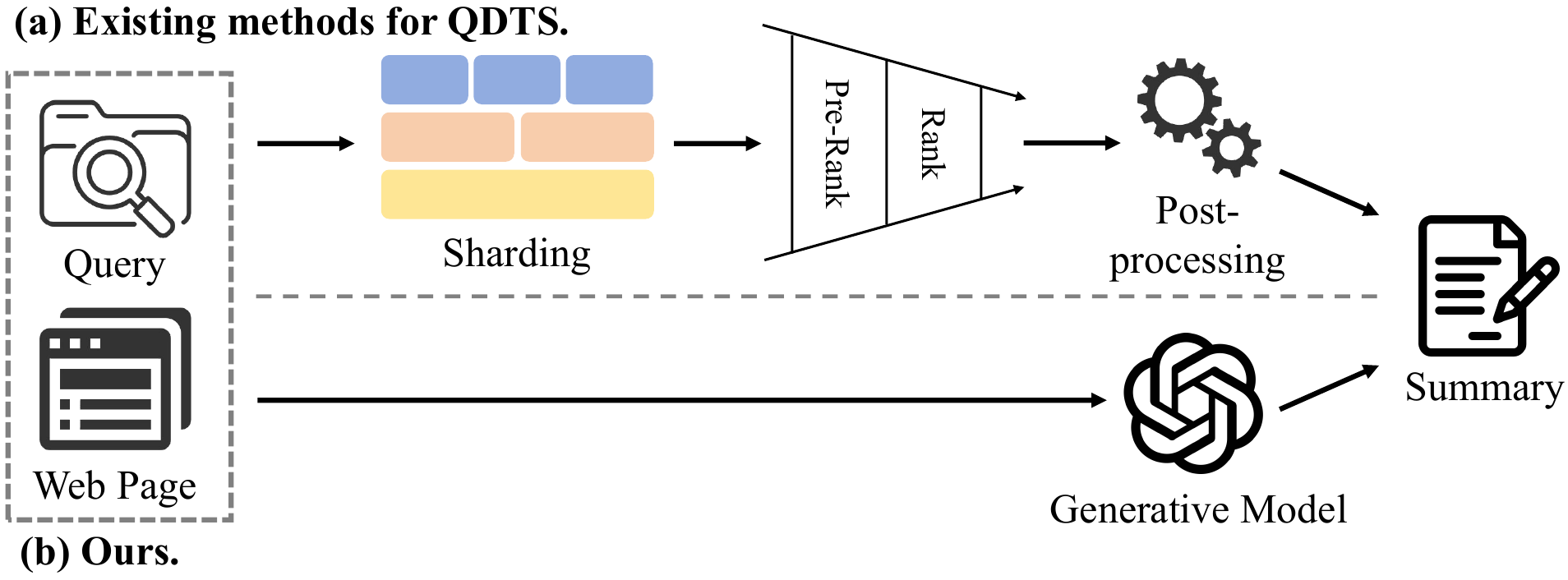} 
  \vspace{-15pt}
  \caption{Differences between existing methods and ours.}

  \label{fig:1}
  % \Description{Task Fig.}
  \vspace{-15pt}
\end{figure}

As shown in Figure \ref{fig:1} (a), the traditional summary generation process typically follows a multi-stage pipeline: extracting content from the landing page, generating snippet candidates, applying pre-ranking, ranking, and finally post-processing. Each stage makes independent selection decisions based on partial information, forming a funnel-like filtering mechanism. This design often leads to the early elimination of relevant content, resulting in cumulative information loss throughout the pipeline~\cite{xu2023lightweight}. Moreover, due to its fragmented structure, such a pipeline is inherently difficult to optimize for the global objective of user intent alignment. Existing summarization methods usually rely on computing the relevance score between the query and the document, selecting the most relevant sentence or paragraph as the summary~\cite{bast2014efficient, turpin2007fast, mohamed2024sdbqfsum}. Many of these systems adopt a "top-1 paragraph selection and concatenation-based presentation" strategy~\cite{mohamed2015similarity, luo2013exploiting}, where a single highly relevant paragraph is selected and then truncated or concatenated with other sentences for display as needed. Due to the lack of global semantic modeling and structural understanding capabilities, this strategy shows clear limitations when dealing with complex real-world search scenarios such as multi-question, multi-faceted, or procedural queries: the model cannot extract relevant information across multiple paragraphs, cannot generate structured summaries, and may even produce content that is irrelevant to the query intent. 

While large language models offer promising capabilities for QDTS through their powerful generation and contextual understanding~\cite{huang2023diffusion, qorib2025just, sotudeh2023qontsum}, their effective integration into real-world systems remains a significant challenge. First, large-scale models are often computationally expensive, suffer from slow inference speeds, and produce unstable outputs, making them difficult to deploy under the strict requirements of industrial applications, such as low latency and high reliability~\cite{li2025proactive}. Second, in the search scenario, we prefer the generated summaries to be extractive in nature, meaning that they should faithfully reflect the content of the original document. This ensures that users can quickly locate relevant information after clicking through to the landing page, rather than being presented with entirely rephrased or hallucinated content.

To effectively address these challenges and generate summaries that are both query-relevant and faithfully reflective of the source document, we propose \textbf{Q}uery-\textbf{D}riven \textbf{Gen}erative \textbf{Sum}marization in \textbf{R}eal-\textbf{T}ime (QDGenSumRT), a novel end-to-end summarization framework in industrial scenarios. QDGenSumRT consists of four key stages: model distillation, supervised fine-tuning (SFT), direct preference optimization (DPO), and deployment optimization, involving model quantization and inference optimization. Specifically, we first perform model distillation by leveraging a 10B-parameter large generative model to produce a massive distilled training dataset. This enables us to obtain a highly efficient 0.1B-parameter lightweight model. Next, to improve the quality and relevance of the generated summaries while keeping them consistent with user preferences, we first use high-quality human-annotated data for SFT to enhance factual correctness; we then use online sampled preference data for DPO to make the model closer to the user's preference and content structure. Finally, we use FP8 quantization and develop model using TensorRT-LLM. Coupled with lookahead decoding strategy, these optimizations ensure the system achieves low latency and high throughput in large-scale search environments. 

% \vspace{-5pt}

The main contributions of our work are summarized as follows:
1) we reveal the limitations of query-driven text summarization in industrial search engines, including information loss due to multi-stage pipelines and poor generation results for complex scenes;
2) we propose a novel end-to-end framework, QDGenSumRT, for real-time query-driven summarization, which integrates model distillation, supervised fine-tuning, preference alignment, and inference optimization;
3) through comprehensive offline and online evaluations, we demonstrate that our framework not only enhances the quality of generated summaries but also exhibits strong feasibility for industrial deployment.

% \begin{itemize}
%     \item We reveal the limitations of query-driven text summarization in industrial search engines, including information loss due to multi-stage pipelines and poor generation results for complex scenes.
%     \item We propose a novel end-to-end framework, QDGenSumRT, for real-time query-driven summarization, which integrates model distillation, supervised fine-tuning, and preference alignment.
%     % The framework is designed to generate concise, accurate, and user-aligned summaries while remaining computationally efficient.
%     \item Through comprehensive offline and online evaluations, we demonstrate that our framework not only enhances the quality of generated summaries but also exhibits strong feasibility for Industrial deployment.
% \end{itemize}

% \vspace{-5pt}
\section{Related Work}
Existing text summarization methods can be categorized into two types: extractive summarization and abstractive summarization~\cite{giarelis2023abstractive}. Extractive methods aim to identify and select the most important sentences from a source document to construct a summary that captures the key information, while abstractive methods aim to produce novel sentences by analyzing the semantic content of the source text. In this work, we aim to address the task of presenting users with query-relevant information extracted from a web page, thereby facilitating efficient access to salient content within the page. Given the need to preserve the original wording and structural context of the source document, we adopt an extractive summarization framework. 

Early extractive summarization methods were mainly statistics-based. One of the pioneering statistical methods was proposed by Luhn~\cite{luhn1958automatic}, who utilized word frequency and distributional analysis to automatically extract important sentences. ANES~\cite{brandow1995automatic} adopted the term frequency–inverse document frequency to compute word importance, selecting high-scoring sentences based on word weights and positional features to generate news summaries. Graph-based extractive methods emerged as another framework that enhances summarization flexibility by representing documents as graphs. TextRank~\cite{mihalcea2004textrank} is an influential graph-based ranking model that treats sentences as nodes and sentence similarity as edges. It iteratively computes node importance and selects high-scoring sentences as the summary. 
Similarly, LexRank~\cite{erkan2004lexrank} adopts a PageRank-like algorithm to compute sentence centrality, capturing mutual reinforcement relationships among sentences to assess their relevance. 
With the advent of deep learning, neural network-based approaches have gained prominence in extractive summarization. Khosrow et al.~\cite{kaikhah2004automatic} were among the first to propose a neural summarization model that learns and integrates sentence-level features. More recently, DCDSum~\cite{zhang2024dcdsum} reframes the extractive summarization task as a sentence re-ranking problem using contrastive learning. It employs RoBERTa to encode the document and dynamically adjusts the number of extracted sentences via a learnable selector. DeepSumm~\cite{joshi2023deepsumm} integrates topic modeling with a Seq2Seq network to capture the semantic information of a document and proposes a novel sentence scoring method for extractive summarization.

Query-driven text summarization is a more complex task that aims to extract content aligned with a given query from relevant documents~\cite{rahman2015survey}. Chandu et al. ~\cite{chandu2019extractive} proposed a query-based extractive summarization model by combining similarity metrics with clustering algorithms. Afsharizadeh et al.~\cite{afsharizadeh2018query} extracted and evaluated sentence-level features to identify those containing valuable information. The appearance of generative models has introduced a more flexible paradigm for text summarization. Nema et al.~\cite{nema2017diversity} proposed a generative approach that incorporates query attention and diversity attention mechanisms. QFAS-KE~\cite{goyal2025qfas} enhances summary relevance by integrating keyword extraction and fine-tuning of pre-trained models.
% The generative model-based methods mainly focus on generating novel sentences, which is align with the extractive method required by our task.
The emergence of large language models (LLMs) offers a promising new direction for addressing this challenge. DeepExtract~\cite{onan2024deepextract} leverages GPT-4 to generate sentence embeddings and incorporates hierarchical positional encoding to compute importance scores for the remaining context.  

Unlike prior work, our approach leverages an LLM to perform query-driven extractive summarization. By applying model distillation and two-stage model optimization, we obtain a lightweight local model that meets the requirements of real-time querying.

\begin{figure*}[t]
    \centering
    \includegraphics[width=\textwidth]{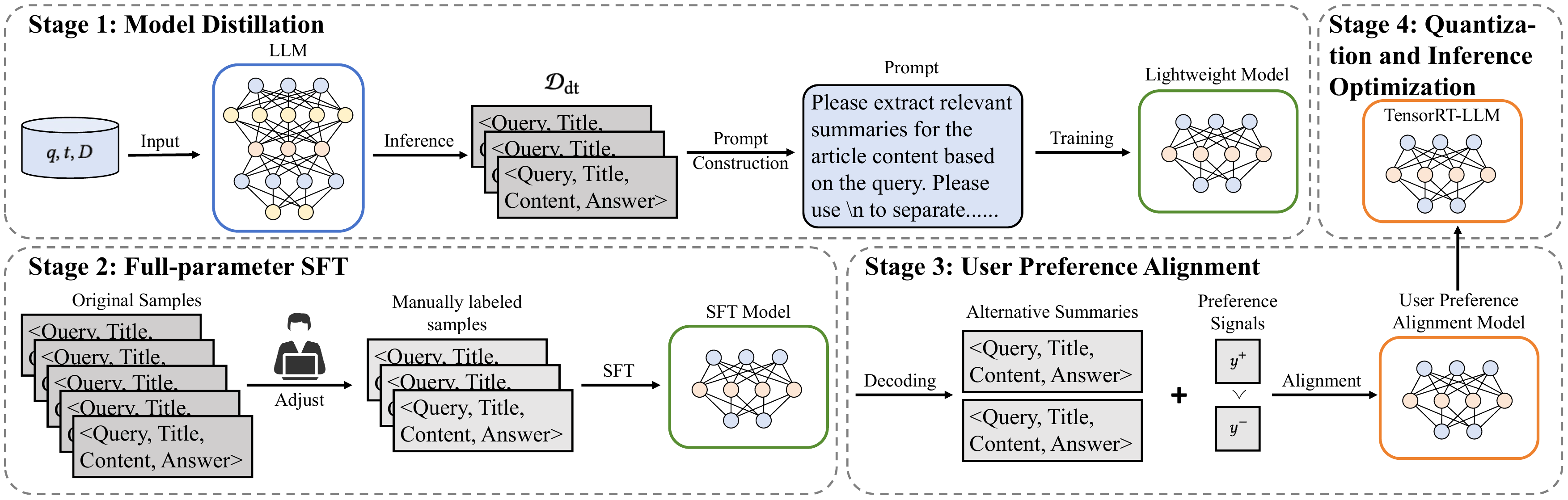}
    \vspace{-15pt}
    \caption{Pipeline overview of QDGenSumRT. We first take user queries, along with web page titles and contents, as input, and employ a teacher LLM to generate a distillation dataset for training a lightweight student model. Subsequently, we fine-tune the student model using a carefully curated human-annotated dataset to enhance the quality and factuality of the generated summaries. Next, we incorporate sampled user preference signals and apply DPO to further align summaries with real user needs. Finally, we deploy the model through quantization and lookahead decoding to achieve optimal efficiency.}
    \vspace{-10pt}
    \label{fig:2}
\end{figure*}

\section{Methodology}

We propose an end-to-end summarization framework QDGenSumRT, as illustrated in Figure \ref{fig:2}. The framework consists of: (1) model distillation, (2) full-parameter supervised fine-tuning, (3) user preference alignment, and (4) model quantization, deployment, and inference optimization. It is designed to simultaneously meet the industrial requirements of high relevance, high quality, and low latency in summary generation. In the first stage, we leverage a 10B-parameter generative model to produce high-quality synthetic training data. These distilled samples are used to train a lightweight student model, effectively compressing the language understanding and extraction capabilities of the large model into a smaller 0.1B-parameter model. In the second stage, we fine-tune the distilled model using carefully curated human-labeled samples. This supervised training further enhances the model’s alignment with the input query and improves the overall relevance of the extracted summaries. In the third stage, to better reflect user preferences in real-world usage, we apply DPO using implicit feedback collected from online interactions. This step encourages the model to generate summaries that are not only relevant but also more likely to satisfy user intent. Finally, to satisfy the real-time requirements of online services, we perform model quantization and deploy the trained model in a high-performance serving environment. Additionally, we adopt a lookahead speculative decoding strategy to accelerate the inference process, enabling large-scale, low-latency generation in production scenarios.

\subsection{Preliminaries}
Query-driven text summarization refers to the task of extracting content from a source document that is relevant to a user-specified query. 
% Each document may be associated with multiple queries, each targeting different aspects or pieces of information within the document.
We formally define the QDTS task as follows.

Given a user query $q$ and a document $D=\{s_1, s_2, ..., s_n\}$, where each $s_i$ denote a sentence, the aim of QDTS is to select a subset of sentences $S=\{s_{o_1}, s_{o_2}, ..., s_{o_k}\}$, such that $S\subseteq D$, $|S| \le k$, and $S$ serves as a query-relevant summary. The selected sentences should not only align with the information need expressed by the query but also preserve essential content from the original document, under a predefined length constraint $k$. Note that each document has a display space constrained by word count in the web search, and the variable $k$ is defined to meet this constraint. In our settings, $k$ is generally set to 80 tokens. We denote our summarization model as a function $f$, such that
\begin{equation}
    S = f(q, D),
\end{equation}
where $f$ takes the query $q$ and document $D$ as input, and outputs the extracted sentence subset $S$.

\subsection{Model Distillation}
Generative language models inherently possess strong capabilities in language understanding and summarization. However, as the number of model parameters increases (e.g., to the scale of 7B or 10B), the inference latency for generating an average of 80 tokens typically exceeds 1000 milliseconds~\cite{grattafiori2024llama, sun2021ernie}, which far surpasses the latency constraints in our industrial scenarios. Although the scaling law of language models suggests that larger models generally outperform smaller ones in terms of generalization and performance, in domain-specific and task-specific settings, well-tuned lightweight models can achieve comparable effectiveness~\cite{kaplan2020scaling}. Moreover, these smaller models often offer better cost-efficiency while satisfying the requirements for real-time inference. To retain the strong representation capabilities of LLM while meeting the requirements of real-time inference, we adopt a teacher–student model distillation approach. Specifically, we use the 10B-parameter ERNIE-Lite-8K model as the teacher and a lightweight 0.1B-parameter Hamburger model as the student. The Hamburger model is built upon the GPT-2 architecture~\cite{radford2019language}, with a hidden size of 768 and 12 layers of self-attention. Through this distillation process, the summarization ability of the large model is effectively transferred to the smaller model.

\noindent\textbf{Distillation dataset construction.} To construct the training data for model distillation, we collect supervision samples from fine-tuned ERNIE-Lite-8K. Specifically, when a user initiates a search query $q$, a ranked list of web pages $U={url_1, url_2, ..., url_n}$ is retrieved. For each URL, we obtain the corresponding page title $t_i$ and content $D_i$ via the document indexer. These components—query, title, and content—are then concatenated and fed into the teacher model to produce target summaries. This results in a distillation training dataset $\mathcal{D}_{\text{dt}}$ consisting of $N$ samples:
\begin{equation}
\mathcal{D}_{\mathrm{dt}} = \left\{ \langle q_i, t_i, D_i, y_i \rangle \;\middle|\; y_i = \pi_{T}(q_i, t_i, D_i; \theta_T),\; i = 1, \dots, N \right\},
\end{equation}
where $\pi_{T}$ denotes the teacher model and $\theta_T$ its parameters. In total, we generate 14 million training samples using the ERNIE-Lite-8K model.

\noindent\textbf{Prompt design for distillation.} Due to the weaker instruction-following capability of the lightweight student model, we should not design a complex prompt. Instead, we adopt a concise and efficient prompt, which also helps reduce the model's prefilling time.
The prompt is formulated as:
\lstset{
  basicstyle=\ttfamily\small,
  breaklines=true,
  backgroundcolor=\color{gray!10},
  frame=single,
  columns=fullflexible,
  keepspaces=true,
  language=,
}

\begin{lstlisting}
Please extract relevant summaries for the article content based on the query. Please use \n to separate multi-viewpoint and step-by-step summaries
---Query:{query}---Title:{title}---Content:{content}---
\end{lstlisting}

This prompt encourages the model to focus on the semantic alignment between the query and the document content, while also optimizing the input length for reduced token budget and computational cost. By using a clear instruction format and domain-specific phrasing, the prompt improves the effectiveness of knowledge transfer during the distillation process. Finally, we obtained 130B tokens using 14 million samples.

\subsection{Full-parameter Supervised Fine-tuning}

After distillation, the lightweight model acquires basic summarization capabilities. However, in practice, we observe that it still suffers from issues such as low relevance and weak structural consistency in certain scenarios. These issues arise mainly due to distributional biases and inconsistencies in summary length present in the distillation data. To address these limitations, we further perform full-parameter supervised fine-tuning using high-quality, human-annotated samples. 

\noindent\textbf{Human-annotated dataset construction.} To ensure that the generated summaries better align with user expectations, we re-curated the training dataset through manual annotation. Specifically, we adopt targeted strategies to enhance the distribution of positive and negative samples, control summary length, and improve structural consistency. \textbf{First}, we addressed the model’s tendency to generate shallow or loosely relevant summaries, a common issue in distilled models caused by exposure to training samples with mismatched or poorly defined target summaries. During data curation, we removed such ambiguous samples to encourage the model to develop a deeper semantic understanding of the query-document relationship. \textbf{Second}, considering the constraint of inference latency and the effectiveness of the generated summary, the lightweight model must not only accurately locate relevant text spans, but also present a complete and informative summary within the space of three display lines—approximately 80 tokens. Overly short summaries lead to excessive whitespace and insufficient information, while overly long summaries are prone to truncation during inference, resulting in incomplete output and unnecessary computation. Therefore, we removed excessively short samples and truncated overly long ones to enforce consistent summary lengths. \textbf{Third}, for tasks requiring multi-perspective or stepwise summaries, structured outputs are preferred. However, such patterns were sparsely represented and inconsistent in the distillation data, leading to poor model performance on these types. To remedy this, we increased the proportion of structured summary samples and enforced output formatting constraints: summaries were required to contain 3–5 distinct points, with each point controlled in length. This ensures better learning of structured paradigms and improves the model’s ability to generate coherent multi-part answers.

Based on carefully curated human-labeled samples, we perform full-parameter supervised fine-tuning on the lightweight model to further enhance its ability to generate summaries that are well aligned with the input query and of high quality. The input is a structured prompt $x=(q, t, D)$, consisting of the user query $q$, web page title $t$, and document content $D$, while the output is the corresponding summary $y$. We formulate this process as an autoregressive generation task, with the objective of maximizing the conditional probability $p(y|q, t, D)$. Specifically, assuming the summary consists of $T$ tokens, the fine-tuning objective is to minimize the following negative log-likelihood loss:
\begin{equation}
\mathcal{L}_{\mathrm{SFT}}(\theta) = -\mathbb{E}_{(q, t, D, y) \sim \mathcal{D}_{\mathrm{SFT}}} \sum_{i=1}^{T} \log \pi(y_i \mid y_{0:i-1}, q, t, D; \theta),
\end{equation}
where $\mathcal{D}_{\mathrm{SFT}}$ denotes the human-annotated fine-tuning dataset, which contains high-quality query-summary pairs with consistent output formatting. $\pi(\cdot)$ represents the output distribution of the lightweight model, and $\theta$ denotes its parameters.

\subsection{User Preference Alignment}
To further align the model outputs with user preferences and generate summaries that better reflect user expectations, we introduce a Direct Preference Optimization step. This step aims to bring the model outputs closer to the implicit user feedback collected during real-time interactions.

We design a live A/B experiment where, for each query-URL pair, the system presents the user with two alternative summaries $y^+$ and $y^-$ at the same display position. User click behavior is used as a weak signal to infer preferences between the two. Specifically, we periodically sample high-traffic query-URL pairs, prioritizing those with stable user engagement and sufficient exposure. For each selected pair, we generate multiple candidate summaries by applying diverse decoding strategies. To improve the quality of the resulting preference dataset used for DPO training, we only retain interaction instances where user click behavior exhibits clear and consistent preference signals. 
% During data collection, we retain only interactions with clearly distinguishable click patterns to ensure the reliability of inferred preferences.
The resulting preference dataset is constructed as follows:
\begin{equation}
\mathcal{D}_{\mathrm{dpo}}=\{(q_i,t_i,D_i,y_i^+,y_i^-)|i=1,\ldots,N\},
\end{equation}
where $y^+_i \succ y^-_i$ indicates that the user preferred $y^+_i$ over $y^-_i$ under the same query $q_i$, title $t_i$, and document content $D_i$.

We adopt the DPO framework~\cite{rafailov2023direct}, which directly optimizes the model to assign higher probability to the preferred summary $y^+$ without requiring reinforcement learning or reward models. The DPO loss is represented by the following equation:
\begin{align}
\mathcal{L}_{\mathrm{DPO}}(\theta) = 
- \mathbb{E}_{(x, y^+, y^-) \sim \mathcal{D}_{\mathrm{dpo}}} 
[ \log \sigma ( \beta \cdot (\log \pi(y^+ \mid x) \notag \\
- \log \pi(y^- \mid x) - \log \pi_0(y^+ \mid x) + \log \pi_0(y^- \mid x) 
))],
\end{align}
where $x=(q, t, D)$ is the structured input, $\sigma$ denote the sigmoid function, $\pi$ denote the current model , and $\pi_0$ be the SFT model as the reference policy. This objective encourages the model to increase the likelihood of preferred responses relative to less preferred ones, while penalizing large deviations from the SFT baseline $\pi_0$, thus achieving preference alignment without reward models. During model optimization, DPO training is interleaved with periodic SFT refresh cycles to adapt to evolving user preferences while maintaining output quality.

\subsection{Quantization and Inference Optimization}
\noindent\textbf{Quantization.} In order to deploy in real-time search systems, we optimize the inference performance of the user preference-aligned summary model through quantization and an efficient serving framework. The model is deployed using TensorRT-LLM inference framework, which provides high throughput, low latency, and optimized memory usage. To further reduce inference time, we apply FP8 quantization~\cite{micikevicius2022fp8}, compressing model weights and activations from bfloat16 to fp8 (W8A8).
% This technique achieves a good balance between efficiency and accuracy. Under the inference setting of 2000 context length and 80 output tokens, the average response time was shortened from 82 milliseconds to 72 milliseconds, and the latency was reduced by about 12\%. 
% Model deployment ensures the robustness and responsiveness of the summary system under high concurrency and production-scale traffic, effectively bridging the gap between LLM capabilities and industry deployment limitations.

\noindent\textbf{Inference optimization.}
To further accelerate inference, we integrate lookahead speculative decoding~\cite{fu2024break} into our serving pipeline. After deploying the model with TensorRT-LLM for optimized inference, we combine FP8 quantization with the lookahead decoding strategy. In this approach, the lookahead branch generates multiple candidate tokens in parallel, while the verification branch uses the original model to validate the proposed n-gram candidates. This effectively breaks the strict token-by-token constraint of autoregressive generation, enabling significant inference acceleration with reduced computational overhead—particularly under long-context and low-latency scenarios.

\section{Experiments}
To evaluate the effectiveness and practical impact of our proposed QDGenSumRT method in real-world scenarios, we conduct comprehensive offline and online experiments on a large-scale web search system. Our experimental setup operates real-world query traffic and user interactions, allowing us to measure both the quality of generated summaries and their influence on user search experience.
\subsection{Datasets}
We evaluate our model on a real-world dataset specifically constructed for QDTS. The dataset is collected from real-world query logs and corresponding search results in a production search system. Each item in the dataset consists of a query-document pair. To construct the ground-truth summaries, we employ professional annotators via Baidu's in-house annotation platform. Annotators are instructed to generate concise, informative summaries that directly address the information need expressed in the query, based on the content of the retrieved documents. Each summary is carefully reviewed to ensure factual consistency, relevance, and centrality.

Ultimately, our final dataset contains 6,160 (query, document, summary) triplets for SFT training and 1,021 triplets for testing.
\subsection{Evaluation Metircs}
% We adopt multiple metrics to evaluate the performance of our model. Specifically, we use Recall-Oriented Understudy for Gisting Evaluation (ROUGE) and 3-point accuracy to assess the model's effectiveness in offline experiments. In online experiments, we measure the model's performance in the real production environment using The Good vs. Same vs. Bad (GSB), click rate, search activity, and satisfied consumption rate.

% 1.ROUGE

% 2. 3-point acc

% 3. GSB

% 4. click rate, search activity, satisfied consumption rate.

We adopt multiple metrics to evaluate the performance of our model, as follows:
% Specifically, we use Recall-Oriented Understudy for Gisting Evaluation (ROUGE)~\cite{lin2004rouge} and 3-point accuracy to assess the model's effectiveness in offline experiments. In online experiments, we measure the model's performance in the real production environment using The Good vs.\ Same vs.\ Bad (GSB), click rate, search activity, and satisfied consumption rate.

% The \textbf{ROUGE} metric family evaluates the quality of generated responses by measuring n-gram overlaps between the model output and reference texts. We report ROUGE-1, ROUGE-2, and ROUGE-L scores to capture different aspects of recall-oriented evaluation.
\noindent\textbf{ROUGE.} The {ROUGE}~\cite{lin2004rouge} metric evaluates the overlap between n-grams in the generated outputs and reference texts. We utilize ROUGE-N (N={1,2}) to measure the n-gram overlap, which is defined as:

\begin{equation}
    \text{ROUGE-N} = 
    \frac{
        \sum_{S \in \text{RefSumm}} \sum_{\text{gram}_n \in S} \text{Count}_{\text{match}}(\text{gram}_n)
    }{
        \sum_{S \in \text{RefSumm}} \sum_{\text{gram}_n \in S} \text{Count}(\text{gram}_n)
    }, 
\end{equation}
where $\text{Count}_{\text{match}}(\text{gram}_n)$ denotes the maximum number of n-grams co-occurring in the candidate summary and the set of reference summaries, and $\text{Count}(\text{gram}_n)$ represents the total number of n-grams in the reference summaries.

ROUGE-L is a metric that measures the similarity between a candidate summary and a reference summary based on the longest common subsequence. It is defined as follows:

\begin{equation}
\text{ROUGE-L} = \frac{L(X, Y)}{|Y|},
\end{equation}
where $ L(X, Y) $ denotes the length of the longest common subsequence between the candidate sequence $ X $ and the reference sequence $ Y $, and $ |Y| $ is the length of the reference sequence.

% \noindent\textbf{3-point accuracy.} To evaluate the quality rate of the generated summaries, we compute the {3-point accuracy}. The formula is defined as follows:

% \begin{equation}
% \text{3-point Accuracy} = \frac{{n}_{3-point}}{{N}},
% \end{equation}
% where ${n}_{3-point}$ is the number of summaries rated as 3, $N$ is the total number of samples. A summary is rated as 3 if it accurately extracts relevant information from the content based on the user's query. The ratings (0, 1, 2, or 3) are assigned through human evaluation for each generated summary.

\noindent\textbf{GSB.} The Good vs.\ Same vs.\ Bad (GSB) metric is judged by professionally trained annotators. For a user query, we provide the annotators with both System A and System B search results. The annotators independently label each comparison as Good (result A is better than B), Bad (result B is better than A), or Same (they are equally good or bad), based on the quality of the search results and their relevance to the query. To quantify human evaluation, we use a unified metric denoted $\Delta_{\text{GSB}}$, defined as:
\begin{equation}
    \Delta_{\text{GSB}} = \frac{\#\text{Good} - \#\text{Bad}}{\#\text{Good} + \#\text{Same} + \#\text{Bad}}.
\end{equation}

In addition to human evaluation, we also monitored several online metrics to assess the model's impact in real-time environments. These include \emph{click-through rate}, which reflects the proportion of queries that result in clicks; \emph{search engagement}, which indicates users' activity during search; \emph{satisfactory consumption rate}, which measures the percentage of users who find satisfactory results; \emph{distribution ratio}, which represents the proportion of queries for which at least one web page is successfully delivered to the user per search.

\subsection{Implementation Details}
For the teacher model used in the distillation stage, we adopt a 10B-parameter model based on ERNIE-Lite-8K, further fine-tuned using 1,000 human-annotated query-summary pairs. The fine-tuning process is conducted over 3 epochs, with a checkpoint saved every 64 steps. For the student model in distillation, we build a custom 0.1B-parameter model named Hamburger, following the GPT-2 architecture. The model consists of 12 transformer layers with a hidden size of 768. During distillation, the teacher model generates 14 million summaries, forming a dataset of approximately 130B tokens. We train the student model on this dataset for 3 epochs using a batch size of 192 and the AdamW optimizer. In the SFT stage, we manually filter and obtain a high-quality dataset of 6,160 examples. The model is fine-tuned for 6 epochs with a batch size of 128. A linear learning rate scheduler is applied, and a checkpoint is saved every 30 steps. The distillation stage is conducted using three NVIDIA A100 GPUs (40GB each), while the SFT stage is completed using a single A100. During deployment, we conduct inference on a single NVIDIA L20 GPU using the {TensorRT-LLM v0.17.0} framework~\cite{tensorrtllm}. We employ fp8 W8A8 + fp8 kv cache quantization. The three key parameters of the lookahead algorithm --- window size, ngram size, and verification set size --- are set to 4, 6, and 4, respectively.

\subsection{Competitor System}

\begin{table}[t]
\centering
\caption{ROUGE scores (\%) comparison on our test set. “*” denotes non-real-time methods. }
\vspace{-10pt}
\label{tab:rouge-comparison}
\begin{tabular}{lccc}
\toprule
\textbf{Model}  & \textbf{ROUGE-1}   & \textbf{ROUGE-2}  & \textbf{ROUGE-L}   \\ \hline
Base          & 40.64   & 34.03   & 39.78   \\
DeepExtract~\cite{onan2024deepextract}* & 38.13   & 25.28   & 33.51   \\
BART~\cite{lewis2020bart}*             & 34.14   & 22.75   & 30.64   \\

NGS* & \textbf{60.48} & \textit{50.25} & \textbf{58.31} \\ 
\midrule
\textbf{QDGenSumRT (Ours) }          & \textit{59.21}   & \textbf{51.33}   & \textit{57.12}   \\
\bottomrule
\vspace{-15pt}
\end{tabular}
\end{table}

\begin{table}[t]
\centering
\caption{Human evaluation results measured by $\Delta_{\text{GSB}}$.}
\vspace{-10pt}
\label{tab:gsb-comparison}
\begin{tabular}{lcc}
\toprule
\textbf{Comparison}       &  \textbf{$\Delta_{\text{GSB}}$}  &  \textbf{p-value} \\ 
\midrule
QDGenSumRT vs. Base   & \textbf{+20.68}\% & 1.3e-6\\
QDGenSumRT vs. NGS  & -0.12\% & 0.88\\ 
\bottomrule
\vspace{-15pt}
\end{tabular}
\end{table}

To demonstrate the effectiveness of our approach, we compare it against several strong baselines and variants: 
% Due to the high cost of deploying suboptimal models, we limit our online experiments to comparisons with state-of-the-art systems currently used in production
\begin{itemize}
    \item\textbf{Base}: A multi-stage traditional summarization strategy currently deployed online in Baidu Search. Specific details are omitted due to confidentiality constraints.
    % \item\textbf{NGM}: NGM represents the Nearline Generative Model, which utilizes a fine-tuned ERNIE-Lite-8K 10B model for summary generation. It is limited triggered online using a query-doc caching mechanism.
    \item \textbf{NGS}: The Nearline Generative Summarizer (NGS) employs a fine-tuned ERNIE-Lite-8K 10B model for summary generation. It is conditionally triggered online using a query-document caching mechanism, but it cannot meet the requirements for real-time inference.
    \item\textbf{DeepExtract}~\cite{onan2024deepextract}: Leverages GPT-4 to generate sentence embeddings and incorporates hierarchical positional encoding to compute importance scores for content selection. 
    \item\textbf{BART}~\cite{lewis2020bart} : An encoder-decoder Transformer model pre-trained using a denoising objective. It consists of a bidirectional encoder and an autoregressive decoder. 
    \item\textbf{QDGenSumRT}: The real-time generative summarization model proposed in this work. With only 0.1B parameters, it is specifically designed for low-latency online deployment and supports efficient end-to-end summarization conditioned on both query and document.
\end{itemize}

\subsection{Main Results}

% To evaluate the performance of our proposed method, we compare it against several competitive summarization systems.

\noindent\textbf{Offline experimental results.} As shown in Table \ref{tab:rouge-comparison}, our proposed {QDGenSumRT} achieves strong performance across all ROUGE metrics and significantly outperforms most existing methods. Specifically, QDGenSumRT obtains the highest ROUGE-2 score of {51.33}, even surpassing NGS, a much larger fine-tuned ERNIE-Lite-8K 10B model. This demonstrates that our model generates more coherent and grammatically accurate summaries while maintaining relevance to the references.
Despite its small size (only 0.1B parameters), QDGenSumRT closely matches the performance of NGS, mainly attributed to the refined SFT and DPO stages. This makes it particularly suitable for low-latency deployment scenarios where both performance and efficiency are critical.

Moreover, third-party methods such as {DeepExtract} and {BART} underperform compared to the production-level {Base} system, due to a lack of adaptation to real-world data distributions.

\begin{table}[t]
\centering
\caption{Relative improvements (\%) of our QDGenSumRT method over the base model on online A/B testing metrics. CTR: Click-Through Rate; SE: Search Engagement; SCR: Satisfactory Consumption Rate; DR: Distribution Ratio. The p-values for these experimental metrics are less than 0.05.}
\vspace{-10pt}
\label{tab:online-ab-results}
\begin{tabular}{lcccc}
\toprule
\textbf{Method} & \textbf{CTR } & \textbf{SE} & \textbf{SCR} & \textbf{DR }\\
\midrule
Base            & —            & —              & —   & —        \\
\textbf{QDGenSumRT} & \textbf{$+$0.81\%}         & \textbf{$+$0.33\%}       & \textbf{$+$0.62\%}    & 
\textbf{$+$0.54\%}                            \\
\bottomrule
\end{tabular}
\end{table}

\begin{table}[t]
\centering
\caption{Efficiency comparison on L20 GPU. Inference latency and end-to-end (E2E) latency are reported as averages.}
\vspace{-10pt}
% ``—" indicates data not available.
\label{tab:efficiency-comparison}
\begin{tabular}{lcccc}
\toprule
\multirow{2}{*}{\textbf{Model}} & \multirow{2}{*}{\textbf{\#Params}} & \multirow{2}{*}{\textbf{Train Time}} & \multirow{2}{*}{\textbf{Inference} } & \multirow{2}{*}{\textbf{E2E}} \\
                       &                          &        \textbf{(per epoch)}                 &      \textbf{Latency}     &\textbf{Latency}      \\ 
% \multirow{2}{*}{Model} & \multirow{2}{*}{\#Params} & \multirow{2}{*}{Train Time} & \multirow{2}{*}{Inference} & \multirow{2}{*}{E2E} \\
%                        &                          &        (per epoch)          &      Latency               & Latency              \\ 
\midrule
Base       & —          & —               & —         &\textbf{\textasciitilde60 ms  }                          \\
NGS        & 10 B          & 28 h              & 3773 ms         & 3802 ms                              \\
\textbf{Ours}         & \textbf{0.1 B}         & \textbf{0.8 h}            & \textbf{55 ms}       & 78 ms                                 \\ 
\bottomrule
\vspace{-15pt}
\end{tabular}
% \vspace{-10pt}
\end{table}
\noindent\textbf{Human evaluation results.}
The human evaluation results, measured by $\Delta_{\text{GSB}}$, are summarized in Table \ref{tab:gsb-comparison}. When comparing our system against the production-level Base system, we observe a significant positive gain of \textbf{+20.68\%} in $\Delta_{\text{GSB}}$, with a highly significant p-value of $10^{-6}$, indicating that our model produces substantially better search summaries according to professional annotators. 

In comparison to NGS, our model achieves comparable performance with only a minor $-0.12\%$ difference in $\Delta_{\text{GSB}}$. This demonstrates that despite its significantly smaller size (0.1B vs. 10B parameters), our model remains competitive in terms of perceived summary quality. 
% More importantly, our approach offers substantial advantages in inference cost and latency, making it a more practical and scalable solution for real-time, resource-constrained deployment scenarios.
\begin{table*}[t]
\centering

\caption{Ablation study on different inference optimization strategies. Best result in bold.}
\vspace{-10pt}
\label{tab:ablation-inference}
\begin{tabular}{lccccc|cccc}
\toprule
\textbf{Config} & \textbf{Quant.} & \textbf{GPU} & \textbf{$w$} & \textbf{$n$} & \textbf{$v$} & \textbf{QPS$\uparrow$} & \textbf{AR$\uparrow$} & \textbf{InfT$\downarrow$} & \textbf{ROUGE-2$\uparrow$} \\
\midrule
Baseline        & BF16            & A10          & --   & --   & --   & 31.1 & 1.00 & 153.31 & \textbf{51.91} \\
+ INT8-wo. + A10 & INT8           & A10          & --   & --   & --   & 46.3 & 1.00 & 131.25 & 49.84 \\
+ INT8-wo. + A10 + Lookahead(4,6,4) & INT8       & A10 & 4   & 6   & 4   & 54.6 & 3.11 & 68.41 & 49.87 \\
+ INT8-wo. + L20  & INT8        &L20 & --   & --  & --   & 79.3 & 1.00 & 118.49 & 49.79 \\
+ FP8 + L20      & FP8             & L20          & --   & --   & --   & 115.4 & 1.00 & 101.21 & 51.31 \\
+ FP8 + L20 + Lookahead(5,5,5) & FP8             & L20 & 5    & 5    & 5    & \textit{148.9} & 3.56 & \textit{57.16} & \textit{51.35} \\
+ FP8 + L20 + Lookahead(6,6,6) & FP8             & L20 & 6    & 6    & 6    & 147.8 & \textbf{3.98} & 59.72 & 51.28 \\
\textbf{+ FP8 + L20 + Lookahead(4,6,4)} & FP8     & L20 & 4 & 6 & 4 & \textbf{149.7} & \textit{3.64} & \textbf{55.01} & {51.33} \\
\bottomrule
\end{tabular}
\vspace{-5pt}
\end{table*}

% \noindent\textbf{Effect of inference optimization strategies.}
% % int8 -wo fp8 lookahead window size 

\noindent\textbf{Online A/B test results.} 
We conducted a large-scale A/B test to evaluate the real-world performance of our method against the base model. As shown in Table~\ref{tab:online-ab-results}, our QDGenSumRT approach consistently outperforms the baseline across several key online metrics, demonstrating its effectiveness in improving user interaction and satisfaction.

\begin{table}[t]
\centering

\caption{Ablation study of different training components on ROUGE-2 scores. Baseline denotes pretrained 0.1B model with GPT-2 structure.}
\vspace{-10pt}
\label{tab:ablation}
\begin{tabular}{lccc|c}
\toprule
\textbf{Method} & \textbf{Distil.} & \textbf{SFT} & \textbf{DPO} & \textbf{ROUGE-2} \\
\midrule
Baseline &               &              &              & 5.34 \\
+ Distillation       & \checkmark    &              &              & 33.51 \\
+ SFT                &               & \checkmark   &              & 35.24 \\
+ DPO                &               &              & \checkmark   & 21.62 \\
+ Distillation + SFT & \checkmark    & \checkmark   &              & \textit{45.87} \\
+ Distillation + DPO & \checkmark    &              & \checkmark   & 40.48 \\
\textbf{Full method}          & \checkmark    & \checkmark   & \checkmark   & \textbf{51.33} \\
\bottomrule
\vspace{-15pt}
\end{tabular}
\end{table}
% \vspace{-15pt}

\noindent\textbf{Efficiency comparison.} To evaluate the practicality of our model in real-world deployment scenarios, we conduct a comprehensive efficiency comparison between our proposed {QDGenSumRT} and the larger {NGS} model.
% The experiments are conducted on L20 GPU platforms, using {TensorRT-LLM v0.17.0} for optimized deployment.
As shown in Table \ref{tab:efficiency-comparison}, QDGenSumRT achieves significantly better efficiency across all metrics. On L20 GPU, our model reduces training time by more than 35x compared to NGS. In terms of inference, QDGenSumRT exhibits over 68x lower inference latency and end-to-end response time. These results highlight the superior scalability and cost-effectiveness of our approach, especially in real-time summarization applications where low latency and high throughput are essential.

In summary, these results validate that QDGenSumRT significantly improves the quality of generated summaries and achieves a better balance between efficiency and effectiveness, making it well-suited for industrial deployment.

\subsection{Ablation study}
\noindent\textbf{Main ablation.} We conduct ablation experiments to evaluate the effectiveness of each component of our pipeline, \textit{i.e.}, model distillation, SFT, and DPO.
As shown in Table~\ref{tab:ablation}, we observe the following findings: First, the base model without any refinement achieves a ROUGE-2 score of only 5.34, demonstrating that directly applying the pretrained 0.1B model is insufficient to handle QDTS tasks. Adding model distillation improves the performance to 33.51, indicating that distillation helps transfer knowledge from larger teacher models into our compact architecture. Further incorporating SFT significantly boosts the score to 45.87, showing the importance of fine-tuning on high-quality, query-document aligned summarization data. Finally, applying DPO brings an additional improvement of +5.46, highlighting its effectiveness in aligning the model with human preferences for concise and relevant summaries. These results demonstrate that each stage in our training pipeline contributes meaningfully to the final summarization quality.
% 研究模型蒸馏、SFT、DPO对摘要效果的影响，可以拆分多一点组件

\noindent\textbf{Effect of inference optimization strategies.}
We conduct ablation studies to evaluate the impact of various inference optimization techniques, including quantization methods (INT8-weight-only, FP8), GPU hardware (A10 vs. L20), and lookahead parameters (window size $w$, ngram size $n$, verification set size $v$). The results are evaluated using three key metrics: queries per second (QPS), acceptance rate (AR), and average inference time (InfT).

From Table~\ref{tab:ablation-inference}, we observe the following: 1) {FP8 quantization} achieves better performance than INT8-weight-only. This is because FP8 conversion from BF16 introduces less precision loss compared to INT8, while also offering greater efficiency gains. Specifically, FP8 includes quantization of all of weights, activations, and KV cache, leading to improvements in both computational efficiency and memory utilization.
2) Lookahead decoding is a lossless inference acceleration technique that does not degrade model quality, as evidenced by the comparable or even slightly improved ROUGE-2 scores. Moreover, it significantly reduces output latency by enabling speculative token generation, which improves QPS substantially.
3) Higher AR does not always translate to better overall efficiency in lookahead decoding. This is because lookahead introduces additional computation overhead. Therefore, careful tuning of its parameters is crucial for maximizing throughput and minimizing inference time.
4) The best configuration, {FP8 + L20 + Lookahead(4,6,4)}, achieves the highest QPS of 149.7 and the lowest average inference time of 55.01 ms, while maintaining competitive generation quality with a ROUGE-2 score of 51.33. This demonstrates a well-balanced trade-off between speed and accuracy.

% \noindent\textbf{Effect of number of SFT samples.}
% 研究6000条样本是不是饱和

% \noindent\textbf{Impact of parameter scale on inference efficiency.}
% % 研究不同模型大小下的inf tm

\section{Conclusion}
In this work, we introduced a novel generative framework QDGenSumRT for QDTS in large-scale web search. We leverage model distillation, SFT, DPO, and lookahead decoding to build a highly effective and efficient system. Despite its compact size of only 0.1B parameters, the proposed model achieves state-of-the-art performance across multiple industry metrics. Extensive experiments demonstrate the effectiveness and efficiency of our method.

\section{GenAI Usage Disclosure}
% GenAI Usage Disclosure
% \noindent\textbf{GenAI Usage Disclosure}
This work utilized generative AI tools for minor language editing and polishing of sentences to improve clarity and readability. These tools were not used to generate new content, results, interpretations, or any novel intellectual contributions. 
% The authors are fully responsible for the accuracy, integrity, and originality of all content in this paper.

% No sections, figures, tables, or code were generated using GenAI tools. The use of such tools was limited to stylistic improvements and grammatical corrections, akin to the use of conventional writing aids like Grammarly or standard word processors.
% \begin{acks}
% \end{acks}

\bibliographystyle{ACM-Reference-Format}
\balance
\bibliography{sample-base}


\end{document}